\documentclass[conference]{IEEEtran}
\IEEEoverridecommandlockouts

\usepackage{cite}
\usepackage{amsmath,amssymb,amsfonts}
\usepackage{algorithmic}
\usepackage{graphicx}
\usepackage{textcomp}
\usepackage{xcolor}
\usepackage{subfig}
\usepackage{multirow}
\usepackage{amssymb}
\usepackage{url}
\def\BibTeX{{\rm B\kern-.05em{\sc i\kern-.025em b}\kern-.08em
    T\kern-.1667em\lower.7ex\hbox{E}\kern-.125emX}}

\begin{document}

\title{Benchmarking Class Activation Map Methods for Explainable Brain Hemorrhage Classification on Hemorica Dataset}

\author{
Z. Rafati\textsuperscript{a},  
M. Hoseyni\textsuperscript{b},
J. Khoramdel\textsuperscript{c},
and A. Nikoofard\textsuperscript{b}\\\\
\textsuperscript{a}\textit{Faculty of Computer Engineering, K. N. Toosi University of Technology, Tehran, Iran} \\
\textsuperscript{b}\textit{Faculty of Electrical Engineering, K. N. Toosi University of Technology, Tehran, Iran}\\
\textsuperscript{c}\textit{Faculty of Mechanical Engineering, Tarbiat Modares University, Tehran, Iran} \\
\textit{email: t.rafati1381@gmail.com}
}

\maketitle

\begin{abstract}
Explainable Artificial Intelligence (XAI) has become an essential component of medical imaging research, aiming to increase transparency and clinical trust in deep learning models. This study investigates brain hemorrhage diagnosis with a focus on explainability through Class Activation Mapping (CAM) techniques. A pipeline was developed to extract pixel-level segmentation and detection annotations from classification models using nine state-of-the-art CAM algorithms, applied across multiple network stages, and quantitatively evaluated on the Hemorica\cite{hemorica} dataset, which uniquely provides both slice-level labels and high-quality segmentation masks. Metrics including Dice, IoU, and pixel-wise overlap were employed to benchmark CAM variants. Results show that the strongest localization performance occurred at \textbf{stage 5} of EfficientNetV2-S, with HiResCAM yielding the highest bounding-box alignment and AblationCAM achieving the best pixel-level Dice ($\approx 0.57$) and IoU ($\approx 0.40$), representing strong accuracy given that models were trained solely for classification without segmentation supervision. To the best of current knowledge, this is among the first works to quantitatively compare CAM methods for brain hemorrhage detection, establishing a reproducible benchmark and underscoring the potential of XAI-driven pipelines for clinically meaningful AI-assisted diagnosis.
\end{abstract}


\section{Introduction}

Intracranial hemorrhage (ICH) is a life-threatening condition that requires rapid and accurate diagnosis, as delays in treatment can significantly increase morbidity and mortality.\cite{CACERES2012771} Non-contrast head CT is the standard imaging modality for detecting hemorrhages; however, interpretation is often challenging due to subtle intensity differences, variability in shape and size of lesions, and the presence of confounding pathologies.\cite{khan2013accuracy} Deep learning models have demonstrated strong performance in slice-wise hemorrhage detection, but their “black-box” nature limits clinical adoption. In high-stakes scenarios such as ICH diagnosis, automated algorithms alone are insufficient. clinicians require transparent explanations to validate predictions and maintain trust in the system. This has motivated a growing body of research in explainable AI (XAI), where new methods for interpretable deep learning are introduced almost daily.\cite{van2022explainable}

Research on XAI for brain hemorrhage has progressed slowly, as most prior studies were hindered by the lack of high-quality pixel-level annotations.\cite{ramananda2023class} The Hemorica\cite{hemorica} dataset provides a unique opportunity to address this gap. Unlike many widely used datasets such as RSNA\cite{flanders2020construction} or CQ500\cite{chilamkurthy2018deep} that lack precise segmentation masks, Hemorica\cite{hemorica} contains 327 head CT studies with slice-level labels and detailed pixel-wise annotations across multiple hemorrhage subtypes. This makes it particularly well suited not only for evaluating classification performance but also for quantitatively comparing XAI methods against ground-truth segmentation masks.

The original Hemorica\cite{hemorica} study offered valuable benchmark results but did not fully explore the dataset’s potential. Research on XAI for brain hemorrhage has also been constrained by the absence of accurate pixel-level annotations, limiting quantitative evaluations of explainability methods. The Hemorica\cite{hemorica} dataset addresses this gap by providing 327 head CT studies with slice-level labels and precise segmentation masks, enabling both classification benchmarking and systematic evaluation of XAI approaches.

The primary focus of this study is on systematically comparing state-of-the-art Class Activation Mapping (CAM)\cite{DBLP:journals/corr/ZhouKLOT15} techniques for visual explanation in the context of intracranial hemorrhage (ICH). 
While XAI in medical imaging spans both textual\cite{poli2021generation} and visual\cite{fang2024diffexplainer} approaches, visual explanations, particularly CAM-based methods, are most compatible with radiology workflows, as they highlight image regions responsible for a model’s decision. 
Previous evaluations of CAMs for ICH have been limited by small or non-diverse datasets and the absence of reliable segmentation masks\cite{rasoulian2023weakly}. 
By leveraging the detailed annotations in Hemorica\cite{hemorica}, a comprehensive quantitative and qualitative comparison of multiple CAM variants across different network layers is performed, enabling assessment of which methods provide clinically meaningful and spatially coherent explanations.

In parallel with the XAI analysis, the classification task itself is revisited. 
To ensure that CAM comparisons are conducted under strong baselines, classification performance is optimized through controlled experiments, examining the effects of input resolution, augmentation strategies, class imbalance handling, decision thresholds, and backbone architectures.

The contributions of this study are summarized as follows:
\begin{enumerate}
\item Baseline results reported for the Hemorica\cite{hemorica} dataset classification task are revisited. An in-depth analysis of the deep-learning models used in prior work is provided, and training and evaluation procedures are refined to obtain improved classification results.
\item A comparative analysis of state-of-the-art class attention modules (CAMs) is presented. These modules are evaluated both qualitatively and quantitatively on the brain hemorrhage classification task, and consistent evaluation metrics are reported to facilitate reproducibility.
\end{enumerate}

\noindent The remainder of this paper is organized as follows. 
Section~\ref{sec:relatedworks} reviews prior work on brain hemorrhage diagnosis and explainable AI methods. 
Section~\ref{sec:methodology} describes the proposed methodology, including the dataset, CAM variants, and the EfficientNetV2 backbone. 
Section~\ref{sec:experiments} outlines the experimental setup and evaluation protocol. 
Section~\ref{sec:results} presents the experimental results and comparative analyses. 
Finally, Section~\ref{sec:conclusion} concludes the paper and discusses possible directions for future research.

\section{Related Works}\label{sec:relatedworks}
Over the past years, deep learning has become increasingly applied in medical imaging\cite{RAYED2024101504, bioengineering11101034}, and brain hemorrhage diagnosis is no exception \cite{hoseyni2024comprehensive, doi:10.1148/ryai.230296}. 
A parallel line of research has focused on providing explanations for deep-learning model predictions\cite{kohan2025application}. 
In this section, prior studies on brain hemorrhage and explainable AI (XAI) methods relevant to this task are reviewed.

\subsection{Datasets}
A number of datasets have been introduced for intracranial hemorrhage (ICH) research, but they vary widely in size, quality, and availability of annotations, which directly impacts the type of tasks that can be studied.  

The RSNA Intracranial Hemorrhage dataset is among the largest publicly available collections, consisting of over 25,000 head CT scans with slice-level labels indicating the presence and subtype of hemorrhage.\cite{flanders2020construction} Despite its scale, the dataset does not provide pixel-level segmentation masks, which limits its use for precise localization or quantitative evaluation of explainability methods. As a result, it is primarily employed for slice-wise or exam-level classification tasks.  

The PhysioNet ICH dataset\cite{hssayeni2020intracranial} provides segmentation masks in addition to classification labels, which makes it more suitable for localization studies. However, the resolution of the annotations is relatively coarse, and the number of scans is significantly smaller than RSNA. Consequently, while useful for proof-of-concept segmentation tasks, it does not fully support fine-grained evaluations at clinical scale.  

CQ500 contains around 500 high-quality CT studies, carefully annotated at the study and slice level by multiple radiologists.\cite{chilamkurthy2018deep} The dataset is frequently used for benchmarking ICH detection models due to its diversity and expert labeling. However, similar to RSNA, CQ500 lacks segmentation masks, which again limits its applicability to explainability studies where pixel-level ground truth is required.  

Seg-CQ500\cite{spahr2023label} builds upon CQ500 by providing pixel-level annotations for a subset of 51 studies. These segmentation masks enable preliminary localization experiments, but the very limited number of scans reduces its utility for large-scale training and evaluation.  

Other datasets such as INSTANCE\cite{li2023state} and various institutional collections exist, but they are either not publicly available or are restricted to small cohorts, making reproducibility and wide adoption difficult.  

In contrast, the Hemorica\cite{hemorica} dataset includes 327 non-contrast head CT examinations with slice-level labels and precise pixel-level segmentation masks across five hemorrhage subtypes. The presence of segmentation masks enables both robust classification benchmarks and quantitative evaluation of explainability methods such as Class Activation Mapping (CAM), where alignment between saliency maps and ground-truth lesions can be measured. This makes Hemorica\cite{hemorica} a comprehensive resource for advancing brain hemorrhage detection and interpretability research.  

\subsection{Brain Hemorrhage Diagnosis}
Previous research on brain hemorrhage classification and detection has explored a range of deep learning approaches\cite{islam2024systematic}, from early convolutional neural networks (CNNs)\cite{xu2024automatic} to more recent transformer-based architectures\cite{rasoulian2023weakly}. Many studies focus on slice-wise 2D classification to identify hemorrhage-positive slices\cite{hoseyni2024comprehensive}, while others employ 3D volumetric models to capture inter-slice context\cite{ramananda2025label}. Reported methods include ensemble learning\cite{gudadhe2023classification}, multi-view CNNs\cite{cao2024pmmnet}, and attention-based networks\cite{wang2020segmentation}, often trained on datasets such as RSNA, CQ500, and PhysioNet. 

The original Hemorica\cite{hemorica} study provided valuable baseline results but left several methodological aspects unexplored, potentially limiting the reported performance. Specifically, the experiments did not examine the effects of data augmentation strategies, varying input image resolutions, different class-weight configurations in binary cross-entropy loss, or a range of classification thresholds. The study explored a broad set of architectures, including \textbf{EfficientNetV2 (Small and Large), ResNet (18 and 50), DenseNet (121 and 201), and SwinV2 (Tiny and Small)}, yet it only partially investigated scaling within the EfficientNet family. In this work, we focus primarily on \textbf{EfficientNetV2 backbones}, systematically analyzing architectural scaling together with training configurations. Addressing these limitations through controlled experiments yields a more comprehensive and reproducible performance benchmark.

\subsection{Explainable Artificial Intelligence (XAI)}
Explainable Artificial Intelligence (XAI) seeks to make AI models more transparent and interpretable, a critical requirement in medical imaging for clinical trust, accountability, and diagnostic support. Approaches can be based on \textit{hand-crafted features} (e.g., radiomics descriptors\cite{rundo2024image}) or \textit{intrinsically interpretable models}\cite{balci2023intrinsically} such as decision trees. Deep neural networks, however, typically require post hoc methods to generate human-understandable explanations. These vary by \textit{output modality}—ranging from text-based rationales\cite{jang2023explainable} to visual explanations such as heatmaps and bounding boxes\cite{ramananda2025label}—and by design, being either \textit{model-specific} (e.g., Grad-CAM\cite{selvaraju2017grad}) or \textit{model-agnostic}\cite{van2022explainable}. Post hoc model-agnostic methods like LIME\cite{ribeiro2016should} and SHAP\cite{lundberg2017unified} are widely used, but in medical imaging, CAM-based visualizations remain the most common.  

For brain hemorrhage, prior XAI studies have relied almost exclusively on CAM methods to highlight regions influencing classification, typically using datasets such as RSNA or CQ500. These explanations were mainly assessed qualitatively by radiologists, since pixel-level ground truth was lacking\cite{huy2025interactive}. Extensions such as probabilistic CAMs or weakly supervised localization\cite{rasoulian2023weakly} improved visualization but remained limited by small-scale validation.  

The Hemorica dataset\cite{hemorica}, which uniquely provides both slice-level labels and pixel-level segmentation masks, enables the first systematic \textit{quantitative} evaluation of CAM methods for hemorrhage. CAM outputs can now be benchmarked against ground truth using overlap metrics such as Dice and IoU, moving beyond subjective inspection toward reproducible evaluation. This study therefore focuses on \textit{visual}, \textit{model-agnostic} CAM techniques, leveraging Hemorica’s detailed annotations to provide a comprehensive comparison of localization performance and address a key gap in prior literature.

\section{Methodology}\label{sec:methodology}

\subsection{Dataset}
The Hemorica\cite{hemorica} dataset comprises 327 non-contrast head CT examinations collected from multiple institutions, accompanied by both slice-level labels and pixel-level segmentation masks. 
In total, it contains 12{,}067 axial slices, of which 2{,}679 are labeled as hemorrhage-positive and 9{,}388 as hemorrhage-negative. 
Positive slices are further categorized into five hemorrhage subtypes: intracerebral hemorrhage (ICH, 1{,}616 slices), intraventricular hemorrhage (IVH, 471 slices), epidural hemorrhage (EPH, 157 slices), subdural hemorrhage (SDH, 752 slices), and subarachnoid hemorrhage (SAH, 253 slices). 
However, due to the substantial class imbalance among subtypes and the relatively low frequency of some categories, we adopt a binary classification scheme (hemorrhage vs. non-hemorrhage) to maximize data availability for the positive class and improve model robustness. 
The dataset presents several challenges, including a pronounced class imbalance (with negatives comprising over 75\% of slices), variability in hemorrhage size and shape, and differences in CT acquisition parameters across scanners and institutions. 
These factors make Hemorica\cite{hemorica} a realistic yet challenging benchmark for developing clinically relevant detection models.

For model development, the dataset was partitioned at the \textit{patient level} to avoid data leakage across scans from the same subject. 
A stratified split ensured that hemorrhage-positive and negative cases were proportionally represented in each set. 
Specifically, 80\% of the studies (261 patients, 9{,}693 slices) were used for training and 20\% of the studies (66 patients, 2{,}374 slices) were reserved for testing, with no overlap of patient scans between sets. 
This strategy preserves independence between subsets while maintaining adequate representation of hemorrhage subtypes in both training and testing.

\subsection{Class Activation Mapping (CAM) Methods}
Class Activation Mapping (CAM) techniques are widely used for interpreting deep neural networks, especially in medical imaging tasks where explainability is critical. CAM-based methods aim to highlight discriminative image regions that contribute most strongly to a model’s decision. Given an input image and a target class, these methods generate a heatmap by combining the forward activations of selected convolutional layers with information derived from gradients, perturbations, or statistical decompositions. The resulting saliency maps provide insight into what spatial features influence the classifier’s output, enabling both qualitative assessment and quantitative evaluation of model trustworthiness\cite{DBLP:journals/corr/ZhouKLOT15}.

Over time, multiple CAM variants have been proposed to address limitations of the original CAM method. In this work, we considered the following ten representative approaches:

\begin{itemize}
    \item \textbf{GradCAM} \cite{selvaraju2017grad}: Computes class-discriminative localization maps by weighting feature maps with the average gradient of the class score with respect to the activations.
    
    \item \textbf{HiResCAM} \cite{draelos2020use}: This method computes the explanation map by taking the element-wise (Hadamard) product between the feature map activations and the corresponding backpropagated gradients. Unlike Grad-CAM, which averages gradients across channels to obtain a single importance weight per feature map, HiResCAM preserves the full spatial correspondence by performing multiplication at every spatial location. The resulting maps are then summed across channels to form the final class activation map.

    \item \textbf{GradCAM-ElementWise} \cite{pytorchgradcam}: This method extends the Grad-CAM formulation by computing the element-wise (Hadamard) product of activations and gradients for each feature map. A ReLU operation is applied to the result before summing across channels to generate the final activation map.

    \item \textbf{GradCAM++} \cite{chattopadhay2018grad}: 
    A modification of GradCAM that incorporates higher-order gradients (first-, second-, and third-order derivatives of the class score with respect to activations). This weighting scheme assigns spatially varying importance to different pixels of the feature maps, allowing distinct regions within the same class to contribute separately to the final localization map.

    \item \textbf{XGradCAM} \cite{fu2020axiom}: 
    A variant of GradCAM that enforces axiomatic properties in the weighting scheme. The weights are computed as a combination of gradients and normalized activations, so that each spatial location contributes proportionally to both its activation strength and its gradient with respect to the class score. This formulation ensures that activations directly influence the final localization map.
    
    \item \textbf{AblationCAM} \cite{ramaswamy2020ablation}: Estimates feature importance by iteratively ablating (zeroing out) channels and measuring the corresponding drop in the target class score. A batched implementation is used to improve efficiency.
    
    % \item \textbf{ScoreCAM} \cite{}: Perturbs the input image using scaled feature maps and measures the change in class score. Unlike gradient-based methods, ScoreCAM does not rely on backpropagation, making it more stable but computationally expensive.
    
    \item \textbf{EigenCAM} \cite{muhammad2020eigen}: Projects the activation space onto its first principal component, generating class-agnostic saliency maps. Despite lacking explicit class information, it often yields visually meaningful results.
    
    \item \textbf{EigenGradCAM} \cite{pytorchgradcam}: Combines the advantages of EigenCAM and GradCAM by multiplying the first principal component of the activations with the gradient signal, producing cleaner class-discriminative maps.
    
    \item \textbf{LayerCAM} \cite{jiang2021layercam}: A modification of GradCAM in which spatially-varying weights are computed for each activation map. Instead of assigning a single global weight per channel, LayerCAM multiplies each activation value by the corresponding positive gradient at the same spatial location. This pixel-level weighting preserves local variations in gradient response, making it possible to highlight smaller structures when applied to early or intermediate layers with higher spatial resolution.

    % \item \textbf{FullGrad} \cite{}: Computes the gradients of the biases from all over the network and sums them, producing explanations that incorporate both input-level and intermediate contributions.
    
    % \item \textbf{Deep Feature Factorizations} \cite{}: Applies Non-Negative Matrix Factorization (NMF) on 2D activations to decompose them into interpretable components, revealing latent factors influencing predictions.
    
    % \item \textbf{KPCA-CAM} \cite{}: Similar to EigenCAM but employs Kernel PCA instead of standard PCA, allowing nonlinear decompositions that can capture more complex activation structures.
    
    % \item \textbf{FEM} \cite{}: A gradient-free method that binarizes activations using a thresholding rule of the form: activation $>$ mean $+ k \times$ standard deviation, yielding sparse but interpretable masks.
    
    % \item \textbf{ShapleyCAM} \cite{}: Weighs activations using gradient information combined with Hessian-vector products, inspired by Shapley values from cooperative game theory to attribute feature importance.
    
    % \item \textbf{FinerCAM} \cite{}: Enhances fine-grained classification by comparing similar classes, suppressing shared features, and emphasizing discriminative details that differentiate closely related categories.
\end{itemize}    

Each of these CAM variants emphasizes different aspects of localization accuracy and spatial resolution. By applying them to the last three convolutional layers of the network, it becomes possible to analyze how explanation quality varies across methods and layers, and to determine which approaches provide the most reliable localization of hemorrhage regions.

\subsection{EfficientNet}
EfficientNet is a family of convolutional neural networks that introduced compound scaling to jointly balance depth, width, and input resolution in a principled manner \cite{tan2019efficientnet}, achieving high accuracy with strong computational efficiency. This study employs EfficientNetV2 \cite{tan2021efficientnetv2}, which improves training speed, parameter utilization, and performance on high-resolution inputs through fused-MBConv layers in early stages and MBConv layers in deeper stages. Six variants are evaluated—B0, B1, B2, B3, S, and L—ranging from lightweight models suitable for constrained settings to larger architectures with higher representational capacity. Their comparison under uniform training and evaluation protocols enables systematic assessment of the trade-offs between efficiency and performance for brain hemorrhage detection.

\subsection{Model Architecture}
\begin{figure}[!t]
    \centering
    \includegraphics[width=\linewidth]{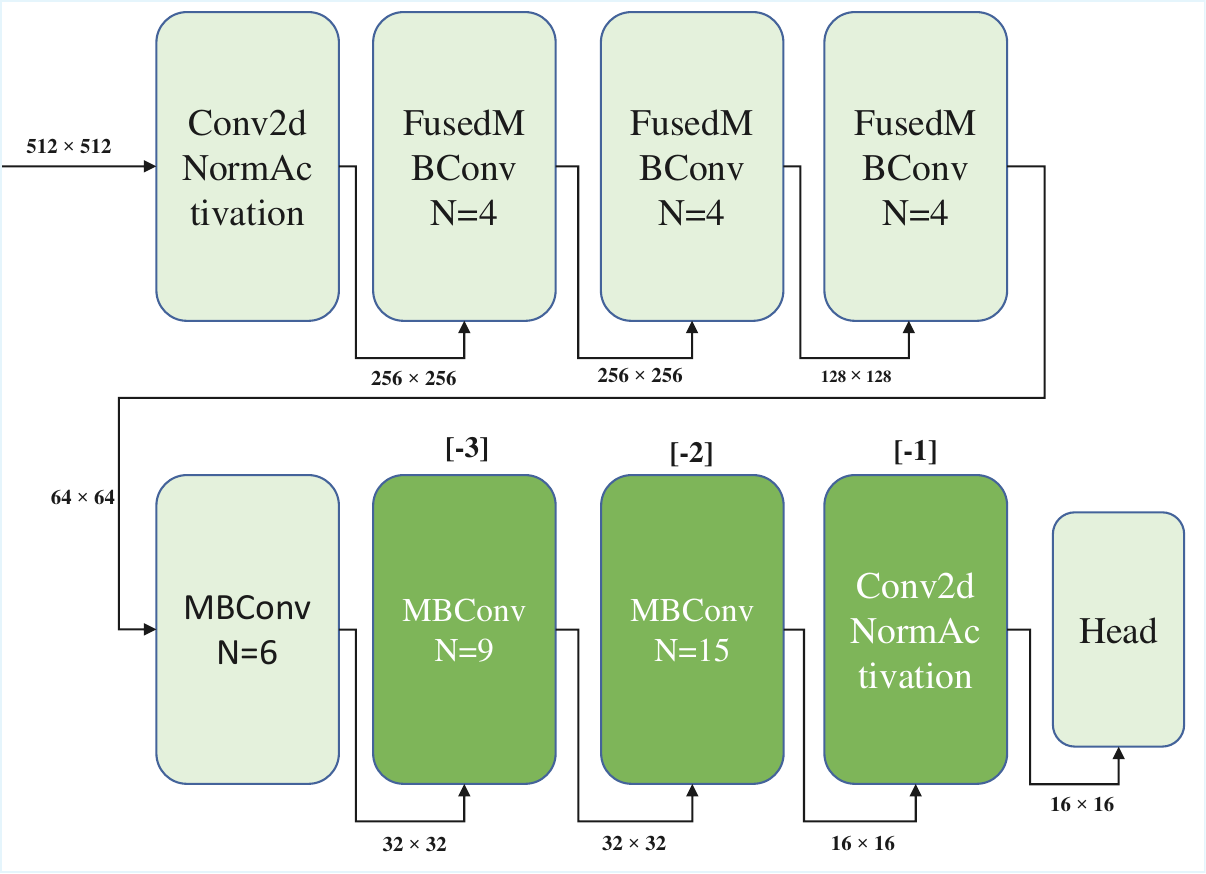}
    \caption{EfficientNetV2-S architecture used in this study. 
    The diagram shows input and output feature map dimensions at each stage. 
    $N$ indicates the number of repetitions of the given block type. 
    The last three convolutional layers ([$-1$], [$-2$], and [$-3$]) correspond to the layers used for CAM-based analysis.}
    \label{fig:effv2s}
\end{figure}

For the classification backbone, EfficientNetV2-S was employed due to its strong balance between accuracy and computational efficiency. 
Figure~\ref{fig:effv2s} illustrates the architecture, showing the progression of feature map dimensions (height $\times$ width) from the input resolution of 512 $\times$ 512 down to the final layers. 
Each stage is composed of repeated building blocks, where $N$ denotes the number of repetitions of that block (e.g., MBConv or FusedMBConv) within the stage. 
The last three convolutional layers, marked as [$-1$], [$-2$], and [$-3$], were selected as the target layers for CAM-based experiments. 

\subsection{Metrics}
The quality of Class Activation Maps (CAMs) was evaluated using pixel-level, bounding-box, and detection metrics. For each slice, CAM heatmaps were thresholded at $\tau$ to produce binary masks $M$, compared against ground-truth masks $G$ from Hemorica\cite{hemorica}.

\subsubsection{Pixel-wise Metrics}
\begin{equation}
\text{IoU} = \frac{|M \cap G|}{|M \cup G|}
\end{equation}

\begin{equation}
\text{Dice} = \frac{2|M \cap G|}{|M|+|G|}
\end{equation}

\subsubsection{Bounding-box Metrics}
To evaluate coarse localization, binary masks were converted into tight bounding boxes around connected regions. The overlap between predicted and ground-truth boxes was then quantified using IoU and Dice at the bounding-box level.

\subsubsection{Loose Hit Rate}
Loose Hit Rate measures whether a CAM mask overlaps the lesion at all. It is assigned a value of 1 if any intersection with the ground truth exists, and 0 otherwise, and is reported as the average across slices.

\subsubsection{Note on True Negatives}
True negative pixels were not included in the evaluation. Since ground-truth masks annotate only hemorrhage regions, background pixels are not explicitly labeled. Metrics such as specificity and accuracy were therefore excluded, and evaluation focused on overlap- and detection-based measures.

\section{Experiments}\label{sec:experiments}

Various factors including model architecture (EfficientNetV2-B0, B1, S, and L), input image resolution, and class imbalance can affect model performance. In this study, experiments were designed to isolate and quantify the influence of these factors on classification results.

Before describing the variations, a \textbf{baseline configuration} was defined and held constant across all experiments unless explicitly changed. The backbone network was EfficientNetV2-B0, trained with a batch size of 16, an initial learning rate of 0.001, and the Adam optimizer for 40 epochs. CT scans were windowed using center 40 and width 80, and no data augmentation was applied. The loss function employed a positive class weight of 1.0, and the input image resolution was set to 224$\times$224. A fixed classification threshold of 0.5 was used to determine positive predictions.

On top of this baseline, several factors were systematically varied to assess their impact on classification performance. Models were trained with input resolutions of $224 \times 224$, $448 \times 448$, and $512 \times 512$ pixels to evaluate how higher spatial detail influences predictive accuracy. To investigate the role of model capacity, multiple EfficientNetV2 variants—including B0, B1, B2, B3, S, and L—were compared, providing insight into the trade-off between performance and computational cost. Class imbalance was addressed by altering the positive weight parameter of the binary cross-entropy loss function, thereby emphasizing hemorrhage-positive slices during training and allowing the effect on sensitivity–specificity balance to be observed. Classification probabilities were binarized at thresholds of 0.3, 0.5, and 0.7 to determine the operating point that maximizes the F1-score. Finally, data augmentation was explored through horizontal flipping ($p=0.5$), rotation (limit $15^\circ$), and shift-scale-rotate transformations (shift limit $0.1$, scale limit $0.1$, rotation limit $15^\circ$). Color jittering was deliberately excluded, as the most distinctive feature for hemorrhage detection in CT images is intensity (grayscale value), and altering it could obscure critical diagnostic cues. These experiments collectively quantified how augmentation contributes to generalization and robustness.

Each experimental variation was compared against a defined baseline configuration: EfficientNetV2-B2, no augmentation, positive weight $=1$, input resolution $224 \times 224$, and classification threshold $0.5$. Sequentially integrating the above factors improved the F1-score of the baseline model by $0.334\%$, achieving a final F1-score of $0.9258$. The corresponding precision, recall, and other evaluation metrics are reported in Table~X.

In the next stage, the reasoning process of the optimized classifier was investigated. Ten widely used Class Activation Mapping (CAM) methods were applied to the last three convolutional layers of the model. For each CAM method and target layer, an empirical optimal threshold was determined and used to generate binary masks from the heatmaps. These CAM-derived masks were then evaluated against ground-truth segmentation annotations from the Hemorica\cite{hemorica} dataset using pixel-wise, bounding-box, and loose-hit metrics, thereby quantifying the alignment of each interpretability technique with actual hemorrhage regions.

\section{Results}\label{sec:results}

This section reports the experimental findings in two parts: (i) classification model performance under different training configurations and EfficientNetV2 backbones, and (ii) quantitative evaluation of CAM-based localization methods.

\subsection{Classification Performance}
Table~\ref{tab:unified_ablation} summarizes the unified ablation study. The baseline configuration, using input size $224 \times 224$, no augmentation, and BCE positive weight of 1, achieved a mean top-5 F1 score of 0.8924. Increasing the input resolution to $448 \times 448$ and $512 \times 512$ consistently improved performance, yielding gains of +0.0135 and +0.0166, respectively. Similarly, introducing data augmentation (random flips, rotations, and scaling) boosted the F1 score to 0.9106, the highest among single-factor ablations. In contrast, modifying the BCE positive weight to 2 or 3 slightly degraded performance, suggesting that the baseline loss weighting already provided a balanced treatment of the hemorrhage and non-hemorrhage classes.

\begin{figure}[htbp]
\centering
\includegraphics[width=0.9\linewidth]{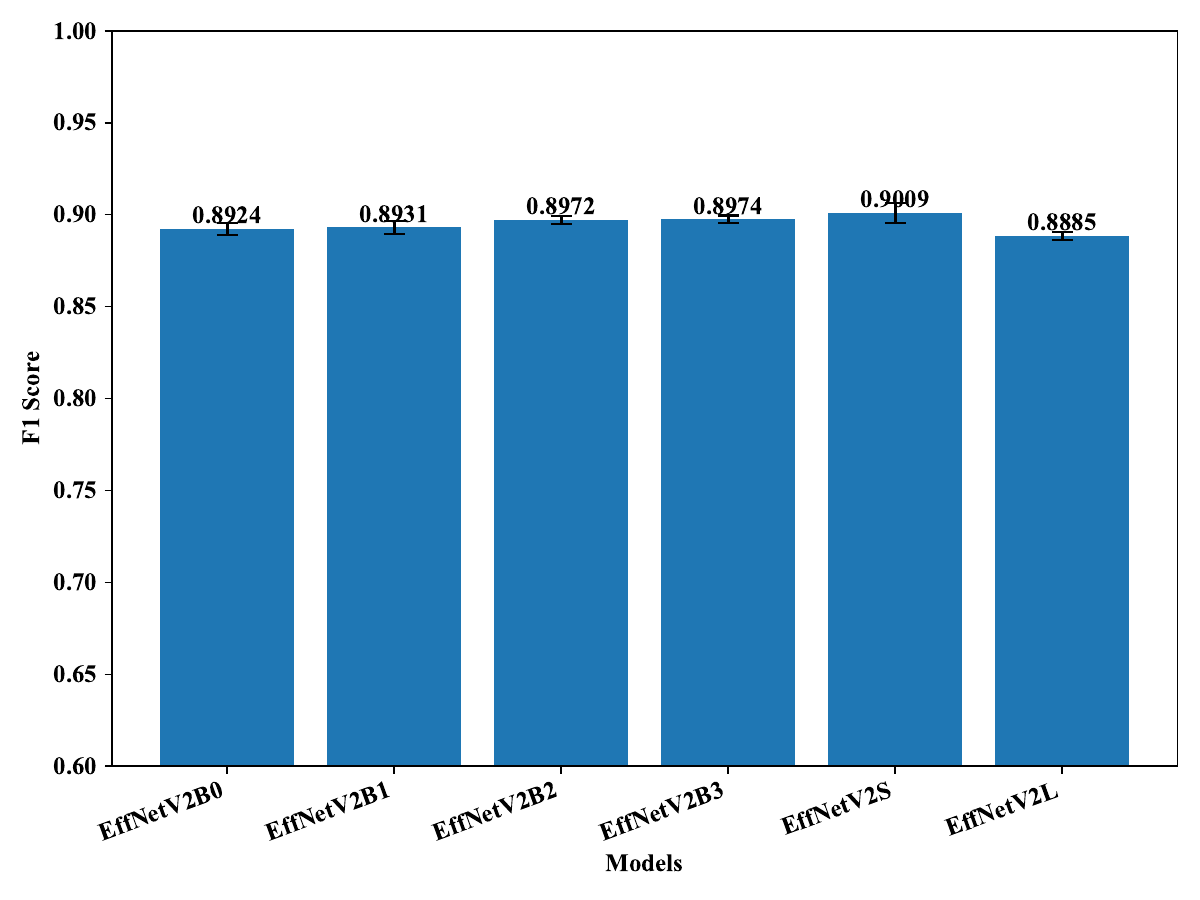}
\caption{Model comparison (mean $\pm$ std of top-5 \textbf{F1 Score}). 
This bar chart compares the performance of different EfficientNetV2 variants under the baseline configuration 
(input resolution 224$\times$224, no augmentation, and BCE positive weight of 1), where the only factor changed is the model architecture. 
Among all evaluated backbones, \textbf{EfficientNetV2-S} achieves the highest F1 score, indicating superior classification performance for this task.}
\label{fig:model_comparison}
\end{figure}

Figure~\ref{fig:model_comparison} compares different EfficientNetV2 architectures under the baseline configuration. EfficientNetV2-S achieved the best performance with a mean top-5 F1 score of 0.9090, outperforming smaller (B0-B3) and larger (L) variants. This indicates that the S variant offers the most favorable trade-off between representational capacity and overfitting for this dataset.

\begin{figure}[htbp]
\centering
\includegraphics[width=0.9\linewidth]{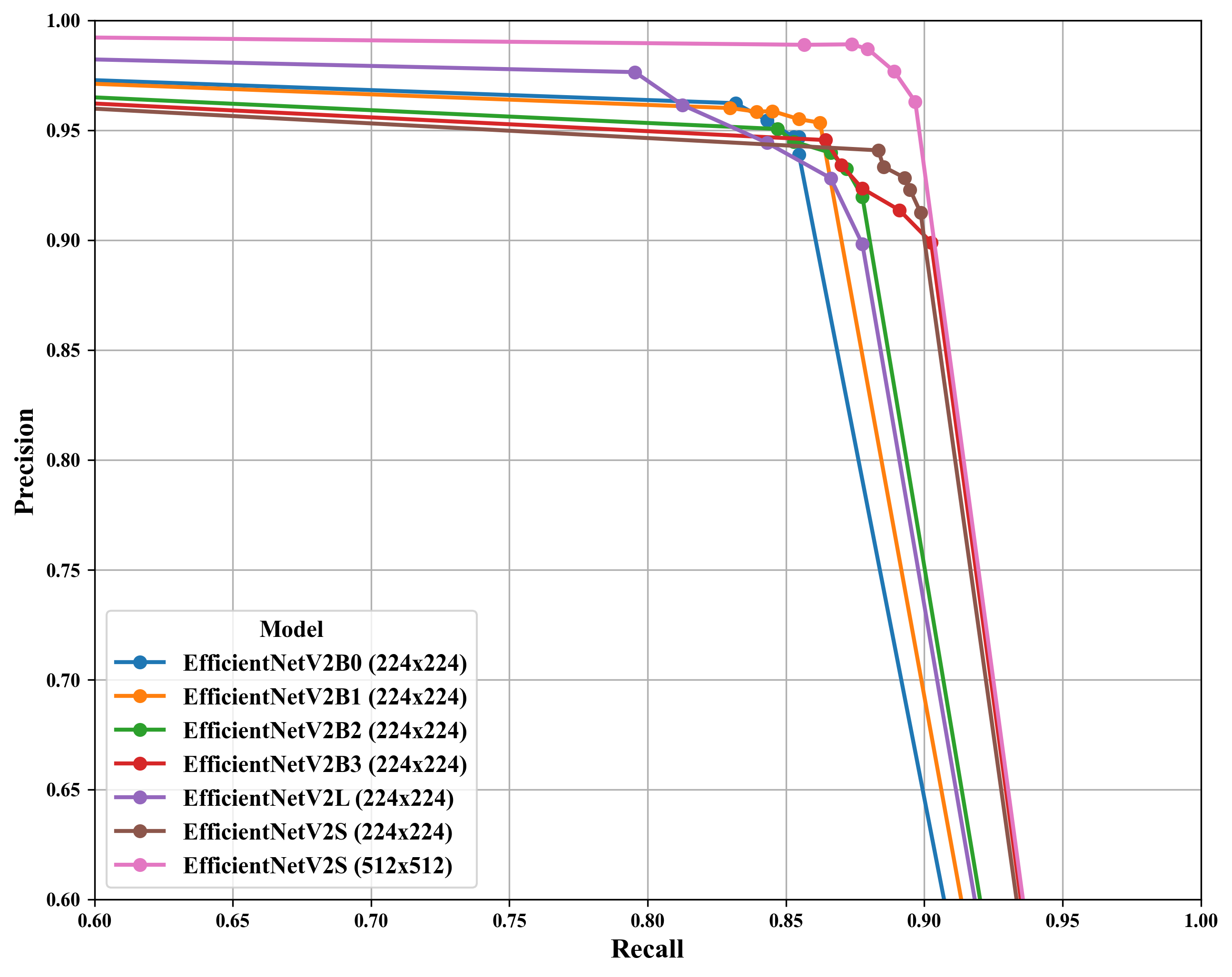}
\caption{Precision-Recall curves of EfficientNetV2 variants at the epoch corresponding 
to the best F1 score. Each curve shows the trade-off between precision and recall as the 
classification threshold is varied from 0.3 (rightmost point) to 0.7 (leftmost point). 
The results demonstrate that all models are relatively insensitive to small changes in 
threshold, maintaining high precision across a wide recall range. Among the compared 
architectures, \textbf{EfficientNetV2-S (512$\times$512)} (pink curve) achieves the 
highest area under the precision-recall curve (AUC), confirming its superiority as the 
final chosen model.}
\label{fig:pr_curve}
\end{figure}

Figure~\ref{fig:pr_curve} shows the precision-recall (PR) curves across classification thresholds (0.3-0.7). All models demonstrated robustness to threshold variation, with minimal sensitivity to changes in decision boundary. Importantly, EfficientNetV2-S trained at $512 \times 512$ with augmentation achieved the highest area under the PR curve (AUC), confirming it as the final selected model for downstream localization experiments.

\subsection{CAM-based Localization}
Tables~\ref{tab:cam_results_global} report the performance of nine CAM variants across three selected convolutional layers of EfficientNetV2-S. Metrics were computed both globally (aggregated across all slices) and on a per-slice average basis.  

The strongest results across all evaluation metrics were consistently obtained from the $[-3]$ layer, which corresponds to the fifth block of the EfficientNetV2-S architecture (see Fig.~\ref{fig:effv2s}). At this depth, HiResCAM achieved the highest bounding-box overlap, while AblationCAM produced the best pixel-level Dice and IoU scores. These findings indicate that intermediate layers, rather than the final convolutional stage, provide the most informative representations for localization.  

The Loose Hit Rate was consistently high ($>0.95$ for most methods), showing that hemorrhage regions were reliably detected across slices. Importantly, the best-performing CAMs achieved pixel Dice scores around 0.57 and IoU scores around 0.40, which represent strong localization quality given that the models were trained solely for classification without access to segmentation supervision.  

In summary, the most accurate localization was observed at the $[-3]$ layer of EfficientNetV2-S, with HiResCAM and AblationCAM providing the top-performing results depending on the metric. Complete quantitative details are available in Tables~\ref{tab:cam_results_global}.

\begin{table*}[htbp]
\caption{Unified ablation study (mean $\pm$ std of top-5 \textbf{F1 Score}). 
Row~1 shows the \textbf{baseline model} (224$\times$224, no augmentation, BCE weight = 1). 
Subsequent rows vary one factor at a time: input resolution (Rows~2--3), data augmentation (Row~4), and BCE positive weight (Rows~5--6). 
The last column ($\Delta$F1) indicates improvement (green) or degradation (red) relative to the baseline.}
\centering
\renewcommand{\arraystretch}{1.2}
\setlength{\tabcolsep}{6pt}
\begin{tabular}{|l|c|c|c|c|c|}
\hline
\textbf{Experiment} & \textbf{Input Res.} & \textbf{Aug.} & \textbf{Pos. W.} & \textbf{F1 Score (Top-5)} & $\Delta$F1 \\
\hline
\textbf{Baseline} & 224$\times$224 & -- & 1 & 0.8924 $\pm$ 0.0033 & -- \\
\hline
\multirow{2}{*}{Input Resolution}
  & 448$\times$448 & -- & 1 & 0.9059 $\pm$ 0.0018 & \textcolor{green}{+0.0135} \\
  & 512$\times$512 & -- & 1 & 0.9090 $\pm$ 0.0057 & \textcolor{green}{+0.0166} \\
\hline
Augmentation & 224$\times$224 & \checkmark & 1 & 0.9106 $\pm$ 0.0022 & \textcolor{green}{+0.0182} \\
\hline
\multirow{2}{*}{Positive Class Weight}
  & 224$\times$224 & -- & 2 & 0.8910 $\pm$ 0.0018 & \textcolor{red}{-0.0014} \\
  & 224$\times$224 & -- & 3 & 0.8870 $\pm$ 0.0013 & \textcolor{red}{-0.0054} \\
\hline
\end{tabular}
\label{tab:unified_ablation}
\end{table*}

% global metrics
\begin{table*}[htbp]
\centering
\caption{Comparison of nine CAM methods across the last three stages (Stage~7 = [-1], Stage~6 = [-2], Stage~5 = [-3]) of EfficientNetV2-S using localization metrics computed \textbf{globally} across the dataset. Pixel-level metrics (Dice, IoU) measure fine-grained overlap, bounding-box metrics assess coarse alignment, and Loose Hit Rate reflects whether any hemorrhage region was detected. Best results for each metric are shown in \textbf{bold}.}
\renewcommand{\arraystretch}{1.2}
\setlength{\tabcolsep}{4pt}
\begin{tabular}{|l|ccc||ccc||ccc||ccc||ccc|}
\hline
\textbf{CAM Method} & 
\multicolumn{3}{c||}{\textbf{Loose Hit Rate}} & 
\multicolumn{3}{c||}{\textbf{Pixel Dice}} & 
\multicolumn{3}{c||}{\textbf{Pixel IoU}} & 
\multicolumn{3}{c||}{\textbf{BBox Dice}} & 
\multicolumn{3}{c|}{\textbf{BBox IoU}} \\
\hline
 & [-1] & [-2] & [-3] 
 & [-1] & [-2] & [-3] 
 & [-1] & [-2] & [-3] 
 & [-1] & [-2] & [-3] 
 & [-1] & [-2] & [-3] \\
\hline
GradCAM & 
0.9583 & 0.9279 & 0.4573 % Loose Hit Rate [-1,-2,-3]
& 0.3640 & 0.3119 & 0.1739 % Pixel Dice
& 0.2225 & 0.1848 & 0.0953 % Pixel IoU
& 0.4117 & 0.4058 & 0.2154 % BBox Dice
& 0.2592 & 0.2546 & 0.1207 \\ % BBox IoU
\hline
HiResCAM & 
0.9602 & 0.9620 & \textbf{0.9753} % Loose Hit Rate [-1,-2,-3]
& 0.4219 & 0.4422 & 0.5219 % Pixel Dice
& 0.2673 & 0.2839 & 0.3530 % Pixel IoU
& 0.4883 & 0.5176 & \textbf{0.5723} % BBox Dice
& 0.3230 & 0.3492 & \textbf{0.4009} \\ % BBox IoU
\hline
GradCAMElementWise & 
0.9620 & 0.9658 & 0.9734 % Loose Hit Rate [-1,-2,-3]
& 0.4508 & 0.4523 & 0.5117 % Pixel Dice
& 0.2910 & 0.2922 & 0.3438 % Pixel IoU
& 0.5049 & 0.5248 & 0.5568 % BBox Dice
& 0.3377 & 0.3557 & 0.3858 \\ % BBox IoU
\hline
GradCAM++ & 
0.9564 & 0.9279 & 0.3814 % Loose Hit Rate [-1,-2,-3]
& 0.3650 & 0.3066 & 0.0794 % Pixel Dice
& 0.2232 & 0.1811 & 0.0413 % Pixel IoU
& 0.4135 & 0.4206 & 0.1175 % BBox Dice
& 0.2606 & 0.2663 & 0.0624 \\ % BBox IoU
\hline
XGradCAM & 
0.7173 & 0.6357 & 0.5750 % Loose Hit Rate [-1,-2,-3]
& 0.0772 & 0.2202 & 0.2215 % Pixel Dice
& 0.0401 & 0.1238 & 0.1246 % Pixel IoU
& 0.1303 & 0.2736 & 0.2421 % BBox Dice
& 0.0697 & 0.1585 & 0.1377 \\ % BBox IoU
\hline
AblationCAM & 
0.9051 & 0.8994 & 0.8956 % Loose Hit Rate [-1,-2,-3]
& 0.3231 & 0.1065 & \textbf{0.5744} % Pixel Dice
& 0.1927 & 0.0562 & \textbf{0.4029} % Pixel IoU
& 0.3672 & 0.2048 & 0.5679 % BBox Dice
& 0.2249 & 0.1141 & 0.3966 \\ % BBox IoU
\hline
% ScoreCAM          &   &   &   &   &   &   &   &   &   &   &   &   &   &   &   \\
% \hline
EigenCAM & 
0.8482 & 0.9507 & 0.6528 % Loose Hit Rate [-1,-2,-3]
& 0.2244 & 0.4752 & 0.2311 % Pixel Dice
& 0.1264 & 0.3116 & 0.1307 % Pixel IoU
& 0.2197 & 0.5251 & 0.2321 % BBox Dice
& 0.1234 & 0.3561 & 0.1313 \\ % BBox IoU
\hline
EigenGradCAM & 
0.9469 & 0.1822 & 0.8065 % Loose Hit Rate [-1,-2,-3]
& 0.4247 & 0.0117 & 0.2258 % Pixel Dice
& 0.2696 & 0.0059 & 0.1272 % Pixel IoU
& 0.4770 & 0.0357 & 0.2879 % BBox Dice
& 0.3132 & 0.0182 & 0.1681 \\ % BBox IoU
\hline
LayerCAM & 
0.9602 & 0.9696 & 0.9753 % Loose Hit Rate [-1,-2,-3]
& 0.4312 & 0.4398 & 0.4909 % Pixel Dice
& 0.2749 & 0.2818 & 0.3253 % Pixel IoU
& 0.4919 & 0.5277 & 0.5384 % BBox Dice
& 0.3261 & 0.3584 & 0.3684 \\ % BBox IoU
\hline
% FullGrad          &   &   &   &   &   &   &   &   &   &   &   &   &   &   &   \\
% \hline
% Deep Feature Factorizations &   &   &   &   &   &   &   &   &   &   &   &   &   &   &   \\
% \hline
% KPCA-CAM          &   &   &   &   &   &   &   &   &   &   &   &   &   &   &   \\
% \hline
% FEM               &   &   &   &   &   &   &   &   &   &   &   &   &   &   &   \\
% \hline
% ShapleyCAM        &   &   &   &   &   &   &   &   &   &   &   &   &   &   &   \\
% \hline
% FinerCAM          &   &   &   &   &   &   &   &   &   &   &   &   &   &   &   \\
% \hline
\end{tabular}
\label{tab:cam_results_global}
\end{table*}

% average metrics
\begin{table*}[htbp]
\centering
\caption{Comparison of nine CAM methods across the last three stages (Stage~7 = [-1], Stage~6 = [-2], Stage~5 = [-3]) of EfficientNetV2-S using multiple localization metrics \textbf{averaged per slice}. Pixel-level metrics (Dice, IoU) assess mask overlap, while bounding-box metrics assess coarse localization. The best result for each metric is highlighted in \textbf{bold}.}
\renewcommand{\arraystretch}{1.2}
\setlength{\tabcolsep}{4pt}
\begin{tabular}{|l|ccc||ccc||ccc||ccc|}
\hline
\textbf{CAM Method} & 
\multicolumn{3}{c||}{\textbf{Pixel Dice}} & 
\multicolumn{3}{c||}{\textbf{Pixel IoU}} & 
\multicolumn{3}{c||}{\textbf{BBox Dice}} & 
\multicolumn{3}{c|}{\textbf{BBox IoU}} \\
\hline
 & [-1] & [-2] & [-3] 
 & [-1] & [-2] & [-3] 
 & [-1] & [-2] & [-3] 
 & [-1] & [-2] & [-3] \\
\hline
GradCAM &
0.3815 & 0.3217 & 0.1574 % Pixel Dice
& 0.2581 & 0.2088 & 0.1082 % Pixel IoU
& 0.4117 & 0.4058 & 0.2154 % BBox Dice
& 0.2592 & 0.2546 & 0.1207 \\ % BBox IoU
\hline
HiResCAM &
0.3838 & 0.4119 & 0.4701 % Pixel Dice
& 0.2594 & 0.2801 & 0.3326 % Pixel IoU
& 0.4883 & 0.5176 & \textbf{0.5723} % BBox Dice
& 0.3230 & 0.3492 & \textbf{0.4009} \\ % BBox IoU
\hline
GradCAMElementWise &
0.4012 & 0.4172 & 0.4682 % Pixel Dice
& 0.2750 & 0.2842 & 0.3303 % Pixel IoU
& 0.5049 & 0.5248 & 0.5568 % BBox Dice
& 0.3377 & 0.3557 & 0.3858 \\ % BBox IoU
\hline
GradCAM++ &
0.3775 & 0.3495 & 0.0898 % Pixel Dice
& 0.2547 & 0.2308 & 0.0594 % Pixel IoU
& 0.4135 & 0.4206 & 0.1175 % BBox Dice
& 0.2606 & 0.2663 & 0.0624 \\ % BBox IoU
\hline
XGradCAM &
0.2789 & 0.2258 & 0.1875 % Pixel Dice
& 0.1901 & 0.1548 & 0.1291 % Pixel IoU
& 0.1303 & 0.2736 & 0.2421 % BBox Dice
& 0.0697 & 0.1585 & 0.1377 \\ % BBox IoU
\hline
AblationCAM &
0.3529 & 0.0902 & \textbf{0.5043} % Pixel Dice
& 0.2364 & 0.0488 & \textbf{0.3759} % Pixel IoU
& 0.3672 & 0.2048 & 0.5679 % BBox Dice
& 0.2249 & 0.1141 & 0.3966 \\ % BBox IoU
\hline
% ScoreCAM &   &   &   &   &   &   &   &   &   &   &   &   \\
% \hline
EigenCAM &
0.3537 & 0.4537 & 0.3103 % Pixel Dice
& 0.2441 & 0.3219 & 0.2266 % Pixel IoU
& 0.2197 & 0.5251 & 0.2321 % BBox Dice
& 0.1234 & 0.3561 & 0.1313 \\ % BBox IoU
\hline
EigenGradCAM &
0.4069 & 0.0156 & 0.2907 % Pixel Dice
& 0.2812 & 0.0092 & 0.2019 % Pixel IoU
& 0.4770 & 0.0357 & 0.2879 % BBox Dice
& 0.3132 & 0.0182 & 0.1681 \\ % BBox IoU
\hline
LayerCAM &
0.3904 & 0.4083 & 0.4595 % Pixel Dice
& 0.2655 & 0.2763 & 0.3226 % Pixel IoU
& 0.4919 & 0.5277 & 0.5384 % BBox Dice
& 0.3261 & 0.3584 & 0.3684 \\ % BBox IoU
\hline
% FullGrad &   &   &   &   &   &   &   &   &   &   &   &   \\
% \hline
% Deep Feature Factorizations &   &   &   &   &   &   &   &   &   &   &   &   \\
% \hline
% KPCA-CAM &   &   &   &   &   &   &   &   &   &   &   &   \\
% \hline
% FEM &   &   &   &   &   &   &   &   &   &   &   &   \\
% \hline
% ShapleyCAM &   &   &   &   &   &   &   &   &   &   &   &   \\
% \hline
% FinerCAM &   &   &   &   &   &   &   &   &   &   &   &   \\
% \hline
\end{tabular}
\label{tab:cam_results_average}
\end{table*}

\begin{figure*}[htbp]
    \centering
    \subfloat[]{\includegraphics[width=0.13\textwidth]{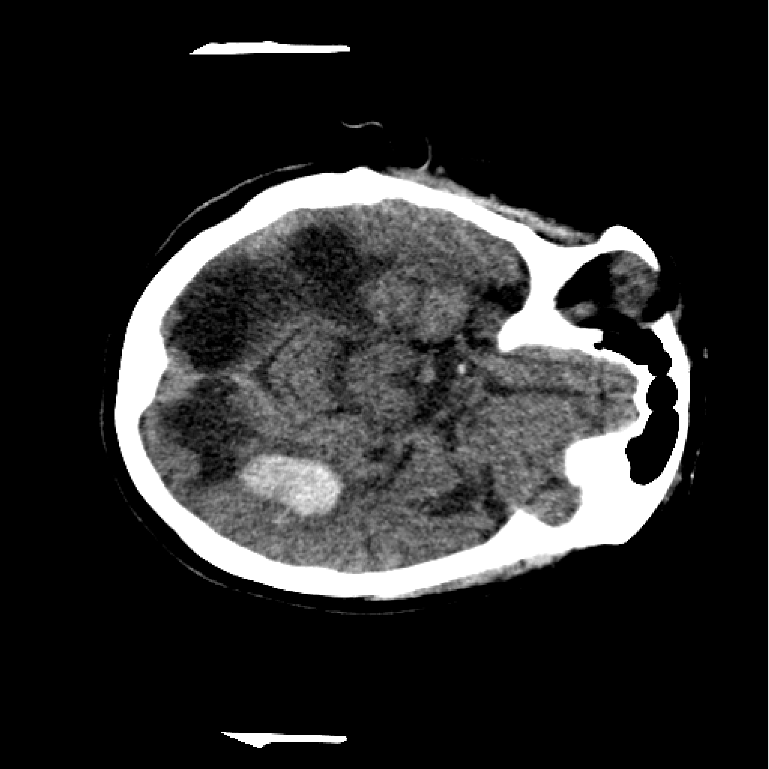}} \hfill
    \subfloat[]{\includegraphics[width=0.13\textwidth]{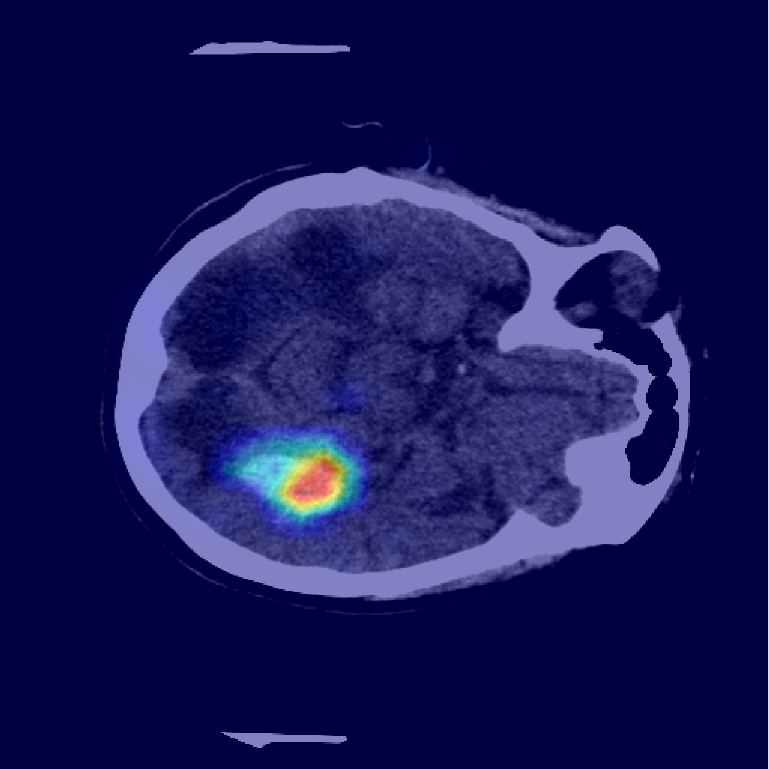}} \hfill
    \subfloat[]{\includegraphics[width=0.13\textwidth]{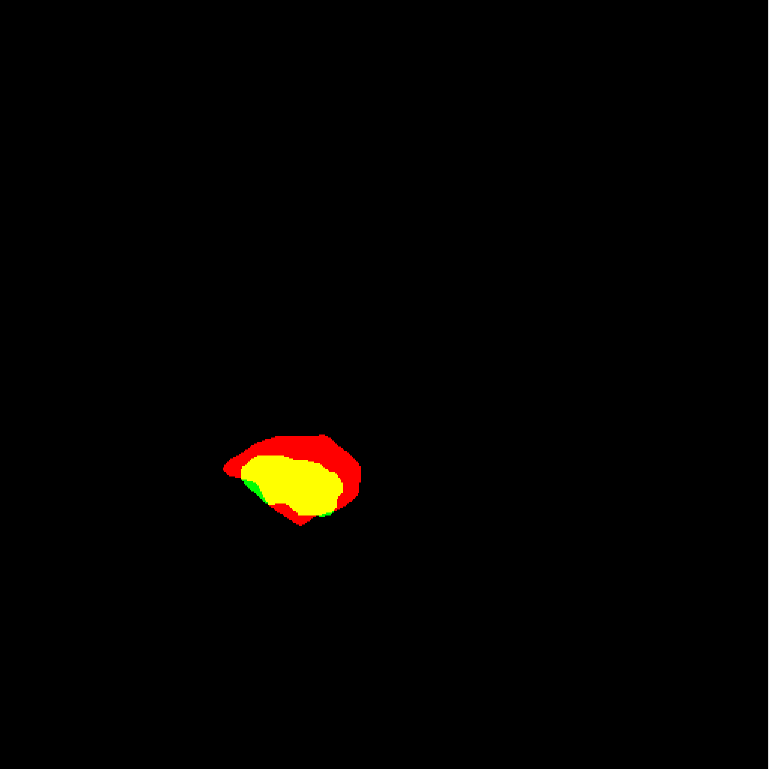}} \hfill
    \subfloat[]{\includegraphics[width=0.13\textwidth]{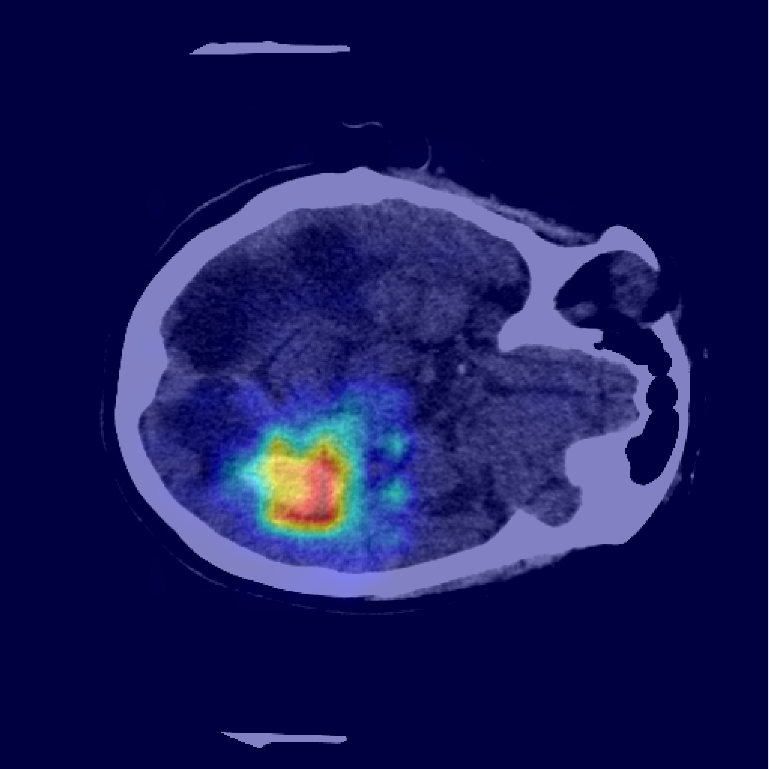}} \hfill
    \subfloat[]{\includegraphics[width=0.13\textwidth]{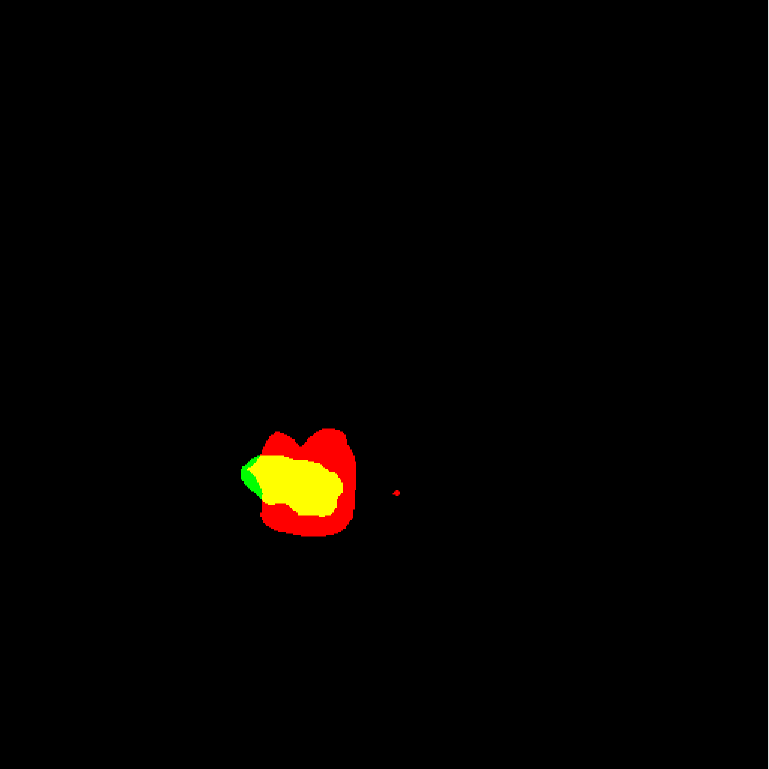}} \hfill
    \subfloat[]{\includegraphics[width=0.13\textwidth]{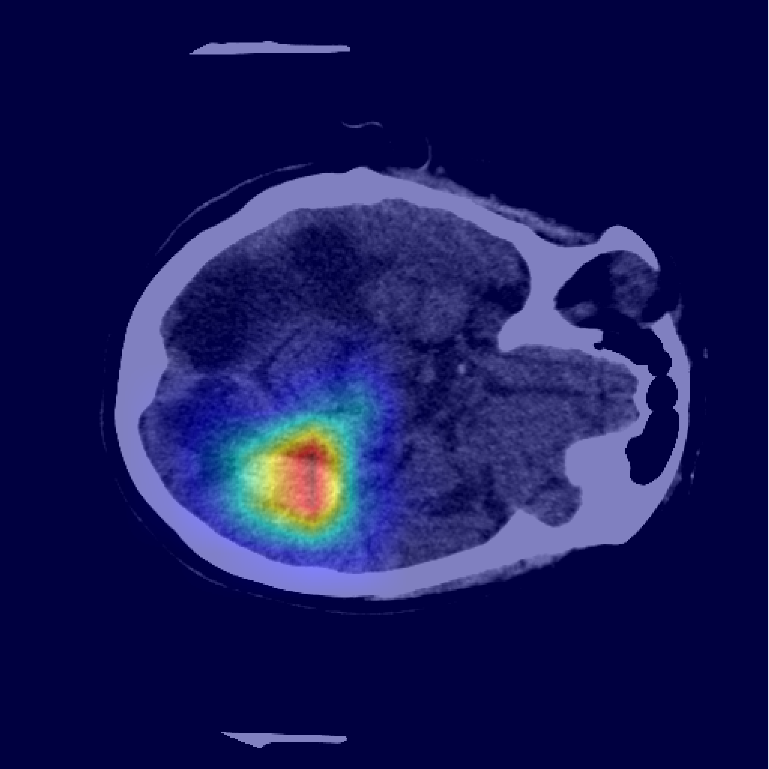}} \hfill
    \subfloat[]{\includegraphics[width=0.13\textwidth]{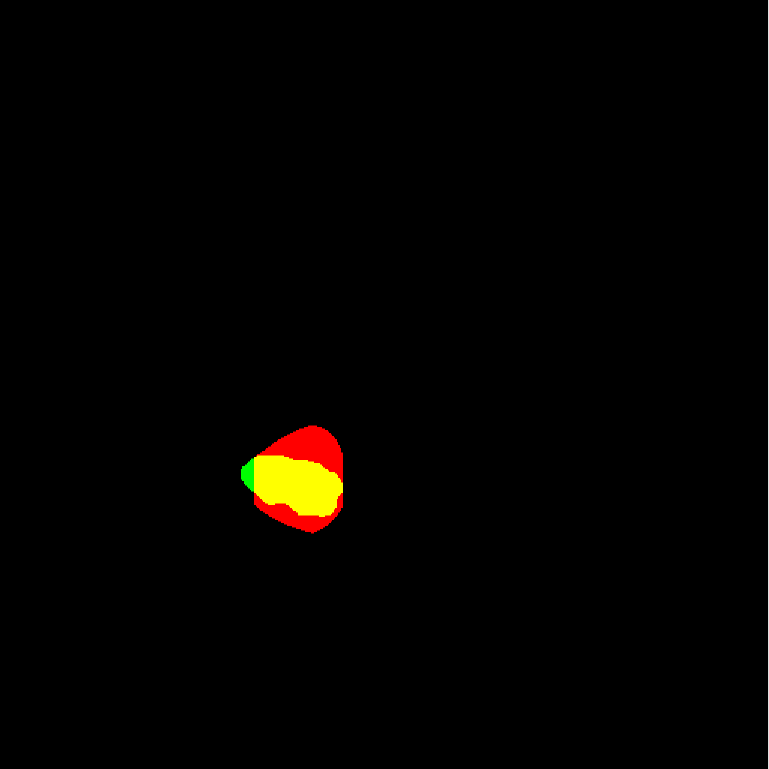}}

    \caption{Example of Class Activation Mapping (CAM) visualizations for a hemorrhage-positive CT slice using HiResCAM across three different network depths.  
    (a) Original CT slice; (b, c) overlay and binary mask from layer $[-3]$; (d, e) overlay and mask from layer $[-2]$; (f, g) overlay and mask from layer $[-1]$.  
    In overlays, the \textcolor{red}{red regions} correspond to CAM-predicted hemorrhage, the \textcolor{green}{green regions} represent the ground-truth segmentation mask, and the \textcolor{yellow}{yellow regions} denote the intersection between prediction and ground truth.  
    This progression illustrates how CAM maps evolve across successive convolutional layers, with earlier layers often capturing broader regions and deeper layers providing more focused localization.}
    \label{fig:cam_example}
\end{figure*}

\section{Conclusion}\label{sec:conclusion}
In this work, a comprehensive study on brain hemorrhage classification and localization was conducted using 
EfficientNetV2 backbones combined with Class Activation Mapping (CAM) techniques. 
Through an extensive set of controlled ablation experiments, input resolution 
and data augmentation were shown to be the most influential factors in improving classification performance, 
while increasing the BCE positive class weight did not yield further benefits. 
Among the tested models, EfficientNetV2-S trained at 512$\times$512 resolution with augmentation 
achieved the best F1 score and the most stable precision-recall behavior across thresholds.

Beyond classification, a systematic comparison of CAM variants was performed across multiple layers 
and evaluation metrics, both globally and averaged per slice. This analysis showed that 
HiResCAM and AblationCAM achieved the strongest localization results in deeper layers of the network. 
By quantifying localization with pixelwise and bounding-box metrics, 
a rigorous benchmark for explainability methods in the context of brain hemorrhage was established.

Overall, the results indicate that careful architectural choices and training configurations, 
combined with interpretable CAM-based visual explanations, can lead to improved classification accuracy 
and clinically meaningful model interpretability. Further improvements may be obtained through 
post-processing techniques designed to suppress spurious activations and refine localization quality.

\bibliographystyle{ieeetr}
\bibliography{mybib.bib}

\begin{thebibliography}{10}

\bibitem{hemorica}
K.~Davoodi, S.~M. Hoseini, J.~Khoramdel, R.~Barati, R.~Mortazavi, A.~H. Nikoofard, M.~Aliyari, and J.~H. Tarikhan, ``Hemorica: A comprehensive ct scan dataset for automated brain hemorrhage classification, segmentation, and detection,'' in {\em arvix}, 2025.

\bibitem{CACERES2012771}
J.~A. Caceres and J.~N. Goldstein, ``Intracranial hemorrhage,'' {\em Emergency Medicine Clinics of North America}, vol.~30, no.~3, pp.~771--794, 2012.
\newblock Acute Ischemic Stroke.

\bibitem{khan2013accuracy}
A.~Khan, S.~Qashqari, and A.-A. Al-Ali, ``Accuracy of non-contrast ct brain interpretation by emergency physicians: A cohort study,'' {\em Pakistan Journal of Medical Sciences}, vol.~29, no.~2, p.~549, 2013.

\bibitem{van2022explainable}
B.~H. Van~der Velden, H.~J. Kuijf, K.~G. Gilhuijs, and M.~A. Viergever, ``Explainable artificial intelligence (xai) in deep learning-based medical image analysis,'' {\em Medical image analysis}, vol.~79, p.~102470, 2022.

\bibitem{ramananda2023class}
S.~H. Ramananda and V.~Sundaresan, ``Class activation map-based weakly supervised hemorrhage segmentation using resnet-lstm in non-contrast computed tomography images,'' {\em arXiv preprint arXiv:2309.16627}, 2023.

\bibitem{flanders2020construction}
A.~E. Flanders, L.~M. Prevedello, G.~Shih, S.~S. Halabi, J.~Kalpathy-Cramer, R.~Ball, J.~T. Mongan, A.~Stein, F.~C. Kitamura, M.~P. Lungren, {\em et~al.}, ``Construction of a machine learning dataset through collaboration: the rsna 2019 brain ct hemorrhage challenge,'' {\em Radiology: Artificial Intelligence}, vol.~2, no.~3, p.~e190211, 2020.

\bibitem{chilamkurthy2018deep}
S.~Chilamkurthy, R.~Ghosh, S.~Tanamala, M.~Biviji, N.~G. Campeau, V.~K. Venugopal, V.~Mahajan, P.~Rao, and P.~Warier, ``Deep learning algorithms for detection of critical findings in head ct scans: a retrospective study,'' {\em The Lancet}, vol.~392, no.~10162, pp.~2388--2396, 2018.

\bibitem{DBLP:journals/corr/ZhouKLOT15}
B.~Zhou, A.~Khosla, {\`{A}}.~Lapedriza, A.~Oliva, and A.~Torralba, ``Learning deep features for discriminative localization,'' {\em CoRR}, vol.~abs/1512.04150, 2015.

\bibitem{poli2021generation}
J.-P. Poli, W.~Ouerdane, and R.~Pierrard, ``Generation of textual explanations in xai: The case of semantic annotation,'' in {\em 2021 IEEE International Conference on Fuzzy Systems (FUZZ-IEEE)}, pp.~1--6, IEEE, 2021.

\bibitem{fang2024diffexplainer}
Y.~Fang, S.~Wu, Z.~Jin, S.~Wang, C.~Xu, S.~Walsh, and G.~Yang, ``Diffexplainer: Unveiling black box models via counterfactual generation,'' in {\em International Conference on Medical Image Computing and Computer-Assisted Intervention}, pp.~208--218, Springer, 2024.

\bibitem{rasoulian2023weakly}
A.~Rasoulian, S.~Salari, and Y.~Xiao, ``Weakly supervised intracranial hemorrhage segmentation using head-wise gradient-infused self-attention maps from a swin transformer in categorical learning,'' {\em arXiv preprint arXiv:2304.04902}, 2023.

\bibitem{RAYED2024101504}
M.~E. Rayed, S.~S. Islam, S.~I. Niha, J.~R. Jim, M.~M. Kabir, and M.~Mridha, ``Deep learning for medical image segmentation: State-of-the-art advancements and challenges,'' {\em Informatics in Medicine Unlocked}, vol.~47, p.~101504, 2024.

\bibitem{bioengineering11101034}
Y.~Xu, R.~Quan, W.~Xu, Y.~Huang, X.~Chen, and F.~Liu, ``Advances in medical image segmentation: A comprehensive review of traditional, deep learning and hybrid approaches,'' {\em Bioengineering}, vol.~11, no.~10, 2024.

\bibitem{hoseyni2024comprehensive}
M.~Hoseyni, K.~Davoodi, F.~Pakdaman, A.~Nikoofard, and M.~Aliyari-Shoorehdeli, ``Comprehensive hyperparameter tuning to enhance deep learning performance for intracranial hemorrhage classification in head ct scans,'' in {\em 2024 31st National and 9th International Iranian Conference on Biomedical Engineering (ICBME)}, pp.~416--423, IEEE, 2024.

\bibitem{doi:10.1148/ryai.230296}
Y.~Wu, M.~Iorga, S.~Badhe, J.~Zhang, D.~R. Cantrell, E.~J. Tanhehco, N.~Szrama, A.~M. Naidech, M.~Drakopoulos, S.~T. Hasan, K.~M. Patel, T.~A. Hijaz, E.~J. Russell, S.~Lalvani, A.~Adate, T.~B. Parrish, A.~K. Katsaggelos, and V.~B. Hill, ``Precise image-level localization of intracranial hemorrhage on head ct scans with deep learning models trained on study-level labels,'' {\em Radiology: Artificial Intelligence}, vol.~6, no.~6, p.~e230296, 2024.
\newblock PMID: 39194400.

\bibitem{kohan2025application}
A.~Kohan, A.~Zahedi, R.~Alizadehsani, R.-S. Tan, and U.~R. Acharya, ``Application of explainable artificial intelligence (xai) techniques in patients with intracranial hemorrhage: A systematic review,'' {\em Wiley Interdisciplinary Reviews: Data Mining and Knowledge Discovery}, vol.~15, no.~3, p.~e70031, 2025.

\bibitem{hssayeni2020intracranial}
M.~D. Hssayeni, M.~S. Croock, A.~D. Salman, H.~F. Al-Khafaji, Z.~A. Yahya, and B.~Ghoraani, ``Intracranial hemorrhage segmentation using a deep convolutional model,'' {\em Data}, vol.~5, no.~1, p.~14, 2020.

\bibitem{spahr2023label}
A.~Spahr, J.~St{\aa}hle, C.~Wang, and M.~Kaijser, ``Label-efficient deep semantic segmentation of intracranial hemorrhages in ct-scans,'' {\em Frontiers in neuroimaging}, vol.~2, p.~1157565, 2023.

\bibitem{li2023state}
X.~Li, G.~Luo, K.~Wang, H.~Wang, J.~Liu, X.~Liang, J.~Jiang, Z.~Song, C.~Zheng, H.~Chi, {\em et~al.}, ``The state-of-the-art 3d anisotropic intracranial hemorrhage segmentation on non-contrast head ct: The instance challenge,'' {\em arXiv preprint arXiv:2301.03281}, 2023.

\bibitem{islam2024systematic}
T.~Islam, M.~S. Hafiz, J.~R. Jim, M.~M. Kabir, and M.~Mridha, ``A systematic review of deep learning data augmentation in medical imaging: Recent advances and future research directions,'' {\em Healthcare Analytics}, vol.~5, p.~100340, 2024.

\bibitem{xu2024automatic}
W.~Xu, Z.~Sha, T.~Tan, W.~Liu, Y.~Chen, Z.~Li, X.~Pan, R.~Jiang, and H.~Yang, ``Automatic segmentation of intracranial hemorrhage in computed tomography scans with convolution neural networks,'' {\em Journal of Medical and Biological Engineering}, vol.~44, no.~4, pp.~575--581, 2024.

\bibitem{ramananda2025label}
S.~H. Ramananda and V.~Sundaresan, ``Label-efficient sequential model-based weakly supervised intracranial hemorrhage segmentation in low-data non-contrast ct imaging,'' {\em Medical Physics}, vol.~52, no.~4, pp.~2123--2144, 2025.

\bibitem{gudadhe2023classification}
S.~S. Gudadhe, A.~D. Thakare, and D.~Oliva, ``Classification of intracranial hemorrhage ct images based on texture analysis using ensemble-based machine learning algorithms: A comparative study,'' {\em Biomedical Signal Processing and Control}, vol.~84, p.~104832, 2023.

\bibitem{cao2024pmmnet}
R.~Cao, D.~Zhang, P.~Wei, Y.~Ding, C.~Zheng, D.~Tan, and C.~Zhou, ``Pmmnet: A dual branch fusion network of point cloud and multi-view for intracranial aneurysm classification and segmentation,'' {\em IEEE Journal of Biomedical and Health Informatics}, 2024.

\bibitem{wang2020segmentation}
J.~L. Wang, H.~Farooq, H.~Zhuang, and A.~K. Ibrahim, ``Segmentation of intracranial hemorrhage using semi-supervised multi-task attention-based u-net,'' {\em Applied Sciences}, vol.~10, no.~9, p.~3297, 2020.

\bibitem{rundo2024image}
L.~Rundo and C.~Militello, ``Image biomarkers and explainable ai: handcrafted features versus deep learned features,'' {\em European Radiology Experimental}, vol.~8, no.~1, p.~130, 2024.

\bibitem{balci2023intrinsically}
A.~T. Balc{\i}, M.~M. Ebeid, P.~V. Benos, D.~Kostka, and M.~Chikina, ``An intrinsically interpretable neural network architecture for sequence-to-function learning,'' {\em Bioinformatics}, vol.~39, no.~Supplement\_1, pp.~i413--i422, 2023.

\bibitem{jang2023explainable}
H.~Jang, S.~Kim, and B.~Yoon, ``An explainable ai (xai) model for text-based patent novelty analysis,'' {\em Expert systems with applications}, vol.~231, p.~120839, 2023.

\bibitem{selvaraju2017grad}
R.~R. Selvaraju, M.~Cogswell, A.~Das, R.~Vedantam, D.~Parikh, and D.~Batra, ``Grad-cam: Visual explanations from deep networks via gradient-based localization,'' in {\em Proceedings of the IEEE international conference on computer vision}, pp.~618--626, 2017.

\bibitem{ribeiro2016should}
M.~T. Ribeiro, S.~Singh, and C.~Guestrin, ``" why should i trust you?" explaining the predictions of any classifier,'' in {\em Proceedings of the 22nd ACM SIGKDD international conference on knowledge discovery and data mining}, pp.~1135--1144, 2016.

\bibitem{lundberg2017unified}
S.~M. Lundberg and S.-I. Lee, ``A unified approach to interpreting model predictions,'' {\em Advances in neural information processing systems}, vol.~30, 2017.

\bibitem{huy2025interactive}
T.~D. Huy, S.~K. Tran, P.~Nguyen, N.~H. Tran, T.~B. Sam, A.~van~den Hengel, Z.~Liao, J.~W. Verjans, M.-S. To, and V.~M.~H. Phan, ``Interactive medical image analysis with concept-based similarity reasoning,'' in {\em Proceedings of the Computer Vision and Pattern Recognition Conference}, pp.~30797--30806, 2025.

\bibitem{draelos2020use}
R.~L. Draelos and L.~Carin, ``Use hirescam instead of grad-cam for faithful explanations of convolutional neural networks,'' {\em arXiv preprint arXiv:2011.08891}, 2020.

\bibitem{pytorchgradcam}
``Pytorchgradcam.'' \url{https://github.com/jacobgil/pytorch-grad-cam}.

\bibitem{chattopadhay2018grad}
A.~Chattopadhay, A.~Sarkar, P.~Howlader, and V.~N. Balasubramanian, ``Grad-cam++: Generalized gradient-based visual explanations for deep convolutional networks,'' in {\em 2018 IEEE winter conference on applications of computer vision (WACV)}, pp.~839--847, IEEE, 2018.

\bibitem{fu2020axiom}
R.~Fu, Q.~Hu, X.~Dong, Y.~Guo, Y.~Gao, and B.~Li, ``Axiom-based grad-cam: Towards accurate visualization and explanation of cnns,'' {\em arXiv preprint arXiv:2008.02312}, 2020.

\bibitem{ramaswamy2020ablation}
H.~G. Ramaswamy {\em et~al.}, ``Ablation-cam: Visual explanations for deep convolutional network via gradient-free localization,'' in {\em proceedings of the IEEE/CVF winter conference on applications of computer vision}, pp.~983--991, 2020.

\bibitem{muhammad2020eigen}
M.~B. Muhammad and M.~Yeasin, ``Eigen-cam: Class activation map using principal components,'' in {\em 2020 international joint conference on neural networks (IJCNN)}, pp.~1--7, IEEE, 2020.

\bibitem{jiang2021layercam}
P.-T. Jiang, C.-B. Zhang, Q.~Hou, M.-M. Cheng, and Y.~Wei, ``Layercam: Exploring hierarchical class activation maps for localization,'' {\em IEEE transactions on image processing}, vol.~30, pp.~5875--5888, 2021.

\bibitem{tan2019efficientnet}
M.~Tan and Q.~Le, ``Efficientnet: Rethinking model scaling for convolutional neural networks,'' in {\em International conference on machine learning}, pp.~6105--6114, PMLR, 2019.

\bibitem{tan2021efficientnetv2}
M.~Tan and Q.~Le, ``Efficientnetv2: Smaller models and faster training,'' in {\em International conference on machine learning}, pp.~10096--10106, PMLR, 2021.

\end{thebibliography}
\end{document}